%% file: arXiv.tex
\documentclass[letterpaper]{article}
\pdfoutput=1  

\usepackage{aaai2027}
\nocopyright

\usepackage[hyphens]{url}
\usepackage{graphicx}
\usepackage{natbib}
\usepackage{caption}
\usepackage{algorithm}
\usepackage{algorithmic}
\usepackage{amsfonts}

\usepackage{newfloat}
\usepackage{listings}
\DeclareCaptionStyle{ruled}{labelfont=normalfont,labelsep=colon,strut=off}
\floatstyle{ruled}
\newfloat{listing}{tb}{lst}{}
\floatname{listing}{Listing}

\usepackage{booktabs}

\usepackage{pifont}
\usepackage{amsmath}
\usepackage{amssymb}
\usepackage{dsfont}
\newcommand{\ind}{\mathds{1}}
\usepackage{colortbl}

\title{
Potential-Guided Particle Steering for Negation-Constrained Dexterous Grasping
}

\author{
    Geonho Kim\equalcontrib,
    SooGon Kim\equalcontrib,
    Jongmin Lee\corresponding
}
\affiliations{
    Department of Computer Science \& Engineering, Chung-Ang University\\
    \{joelkimgh, gon1397, jmlee\}@cau.ac.kr
}

\makeatletter
\let\@oldmaketitle\@maketitle
\renewcommand{\@maketitle}{\@oldmaketitle
  \begin{center}
    \includegraphics[width=0.98\textwidth]{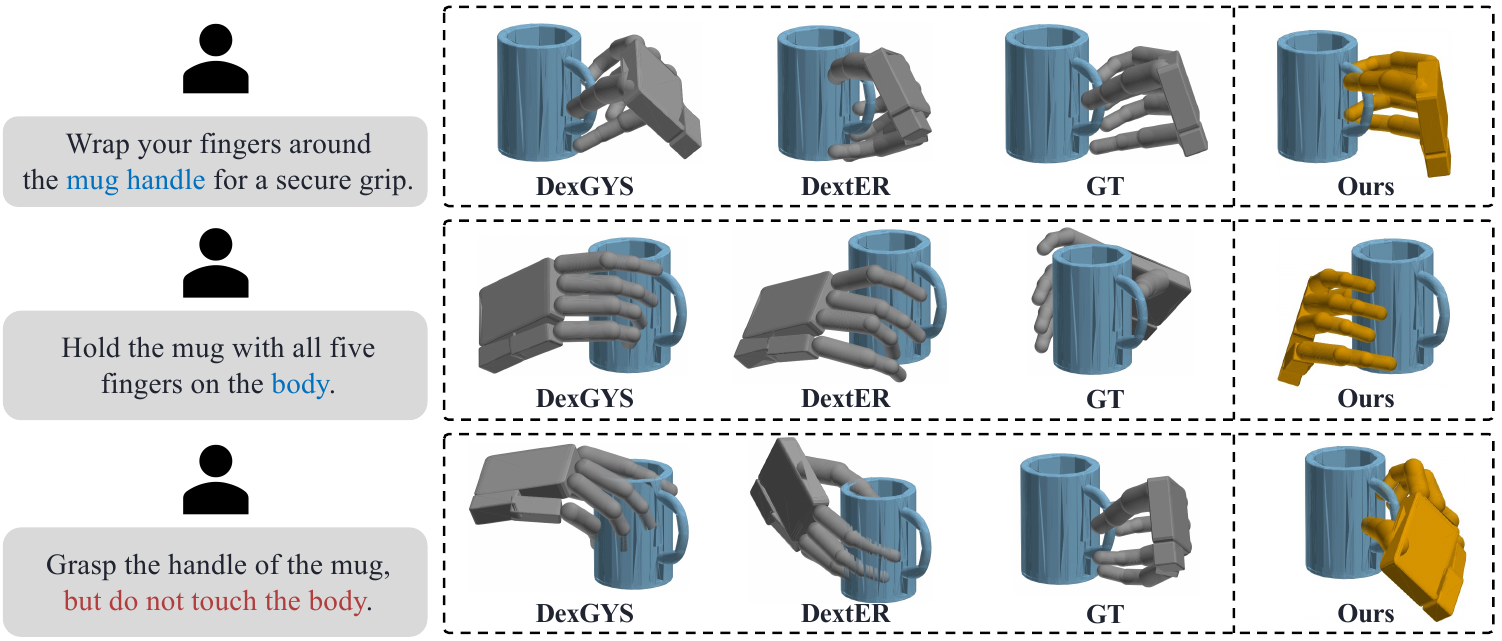}
    \captionof{figure}{\textbf{Grasping under negated instructions.}
Existing models handle positive instructions (top two rows) but drift into the forbidden part when the instruction also says what to avoid (bottom row). Our inference-time steering satisfies both the target and the avoidance constraint in all cases (rightmost column).}
    \label{fig:teaser}
  \end{center}
}
\makeatother

\begin{document}

\maketitle

\begin{abstract}
Language-driven dexterous grasp models, such as DextER, perform well when instructions specify \emph{where} to grasp, but we find they fail systematically when an instruction also specifies where \emph{not} to grasp (e.g., ``grasp the handle but avoid the body''). Existing training corpora, DexGYSNet among them, contain virtually no avoidance instructions, and collecting examples for every possible constraint is impractical. Moreover, because every part mentioned during training denotes a contact target, models may interpret a forbidden part as another region to grasp rather than one to avoid. We therefore introduce an inference-time framework for negation-constrained dexterous grasping that requires no negation-specific training examples. Combining Sequential Monte Carlo with classifier-free guidance, our method guides sampling toward the instructed part while pruning candidates headed for the forbidden region, without any negation examples during training. A frozen 3D part-grounding model localizes the forbidden region from the language instruction. To evaluate this setting, we construct NegGrasp, a benchmark of paired positive/negative instructions with constraint-aware metrics that credit a grasp only if it both accomplishes the task and respects the stated constraint. On NegGrasp, our method reduces the violation rate of the strongest baseline from 57.9\% to 17.2\% while improving both constraint-aware and physical success.
\end{abstract}

\input{Sections/introduction}

\input{Sections/related_works}

\begin{figure*}[t]
\centering
\includegraphics[width=0.98\textwidth]{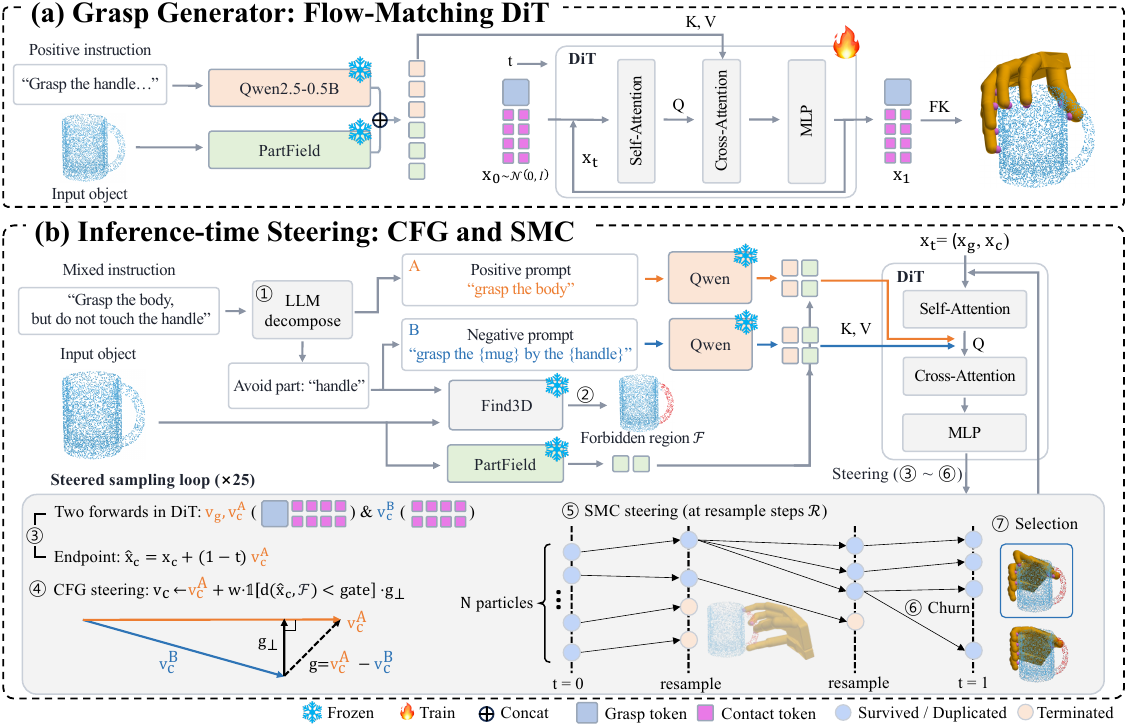}
\caption{\textbf{Pipeline overview.} (a) A trainable DiT denoises the grasp state $x = (x_g, x_c)$ by flow matching, conditioned on the positive instruction and object point cloud via frozen Qwen and PartField encoders. (b) An LLM decomposes the instruction into positive and negative prompts and an avoid part \textcircled{1}, which Find3D grounds into the forbidden region $\mathcal{F}$ \textcircled{2}. Each sampling step forms a look-ahead contact estimate \textcircled{3} and steers it away from $\mathcal{F}$ by contact-gated CFG \textcircled{4}, while SMC resamples particles by constraint compliance \textcircled{5} and churn keeps duplicates from collapsing \textcircled{6}. A final geometric check selects one grasp \textcircled{7}.}
\label{fig:architecture}
\end{figure*}

\input{Sections/method}

\input{Sections/experiments}
\input{Sections/conclusion}


\bibliography{references}

\clearpage
\twocolumn[
  \begin{center}
    {\LARGE \bf Supplementary Material \par}
    \vspace{0.6em}
    {\Large Potential-Guided Particle Steering for Negation-Constrained Dexterous Grasping \par}
    \vspace{1.2em}
  \end{center}
]
\appendix
\input{Sections/appendix_notation}


\end{document}

%% file: Sections/introduction.tex
\section{Introduction}
\label{sec:intro}

Dexterous manipulation with multi-fingered hands is a long-standing goal in robotics, and recent work has shown that natural language can serve as a flexible interface for specifying \emph{how} a robot should grasp an object, such as which fingers to use or which part of the object to contact~\cite{wei2024dexgys,lee2026dexter}. These models, however, are trained and evaluated almost exclusively on \emph{positive} instructions that describe where a grasp should occur. In practice, task and safety constraints are just as often expressed negatively. A user may ask a robot to grasp a knife by the handle and never the blade, or to lift a mug without touching its hot body. Handling such constraints requires a model that can be steered away from disallowed regions while being guided toward a target region. As we show in Figure~\ref{fig:teaser}, current language-driven dexterous grasping models lack this capability.

A generator trained only on positive instructions never observes the association between negation words and avoidance behavior, and so has no reason to acquire it, regardless of its architecture. DexGYSNet~\cite{wei2024dexgys}, the corpus on which both models above are trained, contains essentially no avoidance phrasing (Appendix~\ref{app:datasets}). Indeed, when we give DextER~\cite{lee2026dexter}, our strongest baseline, instructions that pair a positive target with an explicit \emph{negative} clause (e.g., ``grasp the handle of the mug but do not touch the body''), it frequently ignores the clause and contacts the forbidden region (Figure~\ref{fig:teaser}). We further hypothesize that the failure runs deeper than ignoring negation. Because every part name in the DexGYSNet instructions marks somewhere \emph{to} contact, the model learns the shortcut that mentioning a part is itself an instruction to grasp it, so a clause like ``do not touch the body'' is not merely ignored but misread as telling the model where to grasp. This mirrors a broader pattern in vision-language models, which collapse negated and affirmative statements into near-identical representations when never trained to tell them apart~\cite{alhamoud2025negbench}.

A natural fix would be to collect negative instructions and fine-tune on them, which is known to improve negation understanding in vision-language models~\cite{alhamoud2025negbench}. We instead ask whether negation can be handled \emph{without} additional supervision, by steering generation at inference time. Avoidance constraints are user- and context-specific, with a different region forbidden for a different task or user, so enumerating and annotating every constraint in advance is impractical. The natural machinery for such steering is Sequential Monte Carlo (SMC)~\cite{wu2023practical,singhal2025fk}, which prunes and replicates candidates during sampling. What it needs is a mid-generation candidate that predicts the finished grasp yet can still be revised. The contact-token decoder of DextER freezes each token once emitted, and the decoded grasp follows the tokens only approximately (Section~\ref{sec:method-cfg}). A diffusion model instead refines one complete, continuous candidate throughout sampling, so a spatially checkable estimate of the finished grasp is available at every step~\cite{wu2023practical,corso2023particle,skreta2025feynmankac}. We therefore replace the autoregressive grasp decoder with a diffusion transformer (DiT)~\cite{peebles2023dit} trained with flow matching~\cite{lipman2023flow}, and steer its sampling with classifier-free guidance (CFG)~\cite{ho2022classifierfree}, which nudges each candidate away from the forbidden region, and SMC, which prunes candidates already predicted to violate it.

To keep the comparison to DextER controlled, we retain its conditioning backbones, Qwen~\cite{qwen2024report} and PartField~\cite{liu2025partfield}, frozen, and train only the diffusion decoder (Section~\ref{sec:method-arch}). To ground which region a negative instruction refers to, we use Find3D~\cite{ma2025find3d}, an open-vocabulary 3D part-grounding model, as a frozen plug-in during steering.

Our main contributions are:
\begin{itemize}
\item We identify a systematic failure of state-of-the-art language-driven dexterous grasping under negative instructions, and trace it to the extreme scarcity of negation in the training data rather than to a modeling deficiency.
\item We propose an inference-time steering framework that combines CFG and SMC on a flow-matching grasp DiT, together with a marginal-preserving stochastic sampler that keeps resampled particles diverse, enabling avoidance behavior without any negative training examples.
\item We construct NegGrasp, a benchmark of paired positive and negative instructions with constraint-aware metrics for this setting.
\end{itemize}

%% file: Sections/related_works.tex
\section{Related Work}

\subsection{Dexterous Grasp Generation}
Data-driven generative models~\cite{lu2024ugg,weng2024dexdiffuser} trained on large-scale grasp datasets produce physically stable grasps but are agnostic to task intent, which motivates \emph{language-conditioned} grasping. DexGYS~\cite{wei2024dexgys} introduced DexGYSNet, the first large-scale dataset pairing free-form language guidance with dexterous grasp annotations. DextER~\cite{lee2026dexter} builds on this dataset with an autoregressive, contact-token formulation, predicting \emph{which} hand links contact \emph{where} on the object surface before decoding the final grasp. Both lines of work, however, are trained and evaluated purely on instructions that describe a desired contact, and neither targets negative, avoidance-style instructions, which is precisely the gap our work addresses.

\subsection{Negation Prompting}
\subsubsection{Negative prompting in text-to-image diffusion.} Classifier-free guidance (CFG)~\cite{ho2022classifierfree} is the standard mechanism by which text-to-image diffusion models realize \emph{negative} prompts, subtracting the score of an undesired concept from that of the desired one to steer sampling away from it. Follow-up work refines this idea by using perpendicular rather than direct score subtraction, reducing interference between positive and negative guidance~\cite{armandpour2023perpneg}. We adopt this perpendicular formulation, transplanting it from image scores to the contact-coordinate velocities of a grasp generator (Section~\ref{sec:method-cfg}).

\subsubsection{Negation in dexterous grasping.} Two recent works also target avoidance semantics in grasping. OVAL-Grasp~\cite{tong2025ovalgrasp} uses an LLM and a VLM to identify object parts to grasp \emph{or avoid} from a task description, producing a 2D actionable-region heatmap for parallel-jaw, task-oriented grasping. Like our steering, it requires no training, but it operates on image heatmaps for a simple end-effector rather than generating full dexterous hand configurations. AffordDex~\cite{zhao2026afforddex} learns a universal dexterous grasping policy whose reinforcement-learning stage penalizes contact with functionally inappropriate regions via a negative-affordance segmentation module. This bakes in a dataset-level notion of regions that are usually wrong to touch, whereas our constraint is freely specified and changes per instruction at test time, which calls for an inference-time solution rather than a training-time one.

\subsection{Steering Diffusion with SMC and CFG}
Steering methods sample from a reward-tilted version of the model distribution without retraining, and SMC~\cite{wu2023practical,singhal2025fk} achieves this by resampling a population of particles according to importance weights. Each piece of our steering stack (Section~\ref{sec:method-smc}) instantiates a known result from this line. Our potential adopts the max-potential variant of Feynman-Kac steering~\cite{singhal2025fk}, which retains credit for earlier safe checkpoints and is thus robust to noisy intermediate estimates. Feynman-Kac Correctors~\cite{skreta2025feynmankac} show that tilting the drift, as CFG does, and reweighting particles are interchangeable handles of the same framework, licensing our CFG-guided proposal with a corrector on top, while Feynman-Kac-Flow~\cite{mark2025feynmankacflow} extends the framework to conditional flow matching and derives the marginal-preserving sampler that our churn step reduces to.

Other designs in this space make choices we deliberately avoid. The Twisted Diffusion Sampler~\cite{wu2023practical} folds the potential into the proposal via twisted transitions, whereas we keep it as a separate corrector so that the proposal remains the unmodified CFG-guided velocity. Particle Guidance~\cite{corso2023particle} promotes diversity by adding a particle-dependent repulsion term to a coupled reverse SDE with no resampling at all, whereas our churn addresses the same concern through marginal-preserving stochasticity that operates alongside resampling rather than in place of it. Finally, SMC itself is not tied to continuous states, and has steered autoregressive language models with potentials over the partial token sequence~\citep{lew2023sequential}. The contact tokens of DextER make such potentials spatially meaningful, yet our constraint concerns the finished grasp, which the partial sequence neither commits nor can revise once its tokens are emitted, which is why we steer a DiT.

%% file: Sections/method.tex
\section{Method}
\label{sec:method}

Given an object point cloud and an instruction that pairs a positive target with an avoidance clause (e.g., ``grasp the body but do not touch the handle''), our goal is to generate a grasp that reaches the target and never contacts the forbidden region. Because DexGYSNet~\cite{wei2024dexgys} provides no avoidance supervision, we split this into two stages.

Figure~\ref{fig:architecture}a illustrates the first stage, a diffusion transformer (DiT)~\cite{peebles2023dit} over grasp poses and contact coordinates, trained with flow matching~\cite{lipman2023flow} on the positive instructions of DexGYSNet alone and therefore never exposed to a negation. Figure~\ref{fig:architecture}b shows the second, which handles negation entirely at inference time, leaving the trained weights untouched and summarized in Algorithm~\ref{alg:steering}.

\subsection{Grasp Generator: Flow-Matching DiT}
\label{sec:method-arch}

\begin{figure}[t]
\centering
\includegraphics[width=0.9\columnwidth]{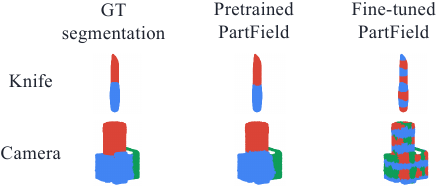}
\caption{\textbf{Encoder erosion under fine-tuning.} Spectral clustering of PartField features into the number of GT parts. Pretrained clusters track GT part boundaries but collapse into fragmented regions after fine-tuning on grasp data.}
\label{fig:partfield-degrade}
\end{figure}

A frozen Qwen~\cite{qwen2024report} encodes the positive-target text into a sequence of hidden states, and a frozen PartField~\cite{liu2025partfield} encodes the object point cloud into a set of patch tokens. We freeze both because fine-tuning erodes the very prior each encoder is there to provide. Qwen supplies the linguistic prior that lets our guidance (Section~\ref{sec:method-cfg}) distinguish a positive mention of a region from a negative one, and fine-tuning large language models on a narrow downstream task has been shown to erode precisely this competence~\cite{luo2023catastrophic}. PartField supplies a geometric prior whose part clustering visibly degraded when we fine-tuned it on grasp data (Figure~\ref{fig:partfield-degrade}). The only trained components are the DiT and a pair of linear layers that project the two conditioning streams into a shared 768-D memory, tagged with a modality embedding.

The state our generator denoises is
\begin{equation}
\label{eq:state}
x = (x_g, x_c), \qquad x_g \in \mathbb{R}^{28}, \quad x_c \in \mathbb{R}^{8 \times 4},
\end{equation}
where $x_g$ is the grasp pose (3D translation, 3D axis-angle rotation, and 22 ShadowHand joint angles) and $x_c$ holds 8 contact slots, each a 3D contact coordinate plus a slot-occupancy flag. A 12-block DiT embeds $x_g$ and the 8 slots as $1{+}8$ tokens (Figure~\ref{fig:architecture}a), and each block applies self-attention among these tokens, then cross-attention to the conditioning memory. The DiT regresses a single velocity field $v_\theta(x, t)$ along the rectified path $x_t = (1-t)x_0 + t x_1$ between noise $x_0 \sim \mathcal{N}(0, I)$ and data $x_1$, so grasp pose and contact tokens are refined together across denoising steps and neither is decoded before the other.


\subsection{Instruction Parsing and Region Grounding}
\label{sec:method-parsing}

Everything the steering needs is derived at inference time, with no access to ground-truth part labels.

\subsubsection{Parsing (Figure~\ref{fig:architecture}b, \ding{192}).} An LLM parser (Gemma 4~\cite{gemma4_2026}) converts a raw instruction into two outputs, a rewrite that describes only the positive target and a part vocabulary of the object in which any part named by a negative word is flagged as forbidden. From these we form the two prompts our guidance contrasts. Prompt A is the positive rewrite and is the only text the generator ever conditions on, while prompt B names the flagged part in the fixed template ``Grasp the \{category\} by the \{part\}'' (e.g., ``Grasp the mug by the handle''). We frame B affirmatively on purpose, since a model trained without negation reads a sentence about what to avoid as one more instruction of where to grasp, the very failure we set out to fix. Appendix~\ref{app:parser} details the parser.

\subsubsection{Grounding (Figure~\ref{fig:architecture}b, \ding{193}).} Find3D~\cite{ma2025find3d} labels every point of the object point cloud with one part name from the parsed vocabulary, and the forbidden region $\mathcal{F}$ is the set of points that receive a flagged name. We hand over the full vocabulary rather than the forbidden part alone because Find3D scores only the closed set of candidates given in a single call. The region never enters the generator and instead constrains sampling from outside the network.

\subsection{Steering by Classifier-Free Guidance}
\label{sec:method-cfg}

Guidance fires only when a candidate is already headed for the forbidden region, and this asks two things of the generator, both properties of how it samples rather than of how well it fits the data. The first is a mid-generation estimate of the finished grasp that the sampler can still revise. A flow model refines one complete, continuous candidate throughout sampling, whereas a contact-token decoder freezes each token once emitted and its grasp follows the tokens only approximately, so even a clean sequence does not certify a clean grasp. Classifier-free guidance itself transfers to autoregressive models~\cite{sanchez2024cfgllm}, but it contrasts the vocabulary of a single token at a time and carries no notion of spatial destination. The second is a handle that admits an immediate geometric check and still moves the grasp. Contact coordinates are points on the object surface, so measuring their distance to the forbidden region needs no extra machinery, while the grasp pose alone would require forward kinematics at every denoising step. Because pose and contacts are refined along the same trajectory, steering the contact velocity reshapes the pose that co-evolves with it.

\subsubsection{Look-ahead estimate (Figure~\ref{fig:architecture}b, \ding{194}).} At each denoising step we form a one-step estimate of the finished contact coordinates,
\begin{equation}
\label{eq:lookahead}
\hat{x}_c = x_c + (1-t)\,v_c^A,
\end{equation}
where $x_c$ is the contact component of the generation state at flow time $t$ and $v_c^A$ that of the velocity predicted under prompt A, before any guidance is applied, so the estimate is read off a forward pass already taken.

\subsubsection{Gated perpendicular guidance (Figure~\ref{fig:architecture}b, \ding{195}).} A second forward pass under prompt B yields $v_c^B$, and we steer the contact velocity by its difference from the positive-conditioned $v_c^A$,
\begin{equation}
\label{eq:guidance}
v_c \leftarrow v_c^A + w \cdot \ind\big[d(\hat{x}_c, \mathcal{F}) < \mathrm{gate}\big] \cdot g_\perp,
\end{equation}
where $g_\perp = \mathrm{orth}(v_c^A - v_c^B,\, v_c^A)$ keeps only the component orthogonal to $v_c^A$, and $w$ is the guidance weight. An unprojected difference risks cancelling the positive target itself, whereas its orthogonal component pushes generation sideways without erasing the target, following perpendicular negative prompting in text-to-image diffusion~\cite{armandpour2023perpneg}. Steering the contact channels alone leaves the pose to conditional generation, and the indicator makes guidance self-limiting, since the term vanishes once a candidate clears the forbidden region and stops perturbing convergence. Algorithm~\ref{alg:steering} additionally exposes a step window $[\mathrm{lo}, \mathrm{hi}]$ that limits guidance in time.

\subsection{Steering by Sequential Monte Carlo}
\label{sec:method-smc}

\subsubsection{A look-ahead potential.} We score the same estimate $\hat{x}_c$ with a potential
\begin{equation}
\label{eq:potential}
\Phi(x_t) = \tfrac{1}{2}\,\psi(\hat{x}_c, \mathcal{F}; \tau) + \tfrac{1}{2}\big(1 - \psi(\hat{x}_c, \mathcal{O}; \delta)\big),
\end{equation}
where $\psi(\cdot, \mathcal{S}; s) = \min\big(\max(d(\cdot, \mathcal{S})/s,\, 0),\, 1\big)$ rises from 0 to 1 as the estimated contacts move away from a set $\mathcal{S}$, $d$ is the minimum Euclidean distance, and $s$ is a fixed saturation distance, $\tau$ for the forbidden region $\mathcal{F}$ and $\delta$ for the object surface $\mathcal{O}$. $\Phi \in [0,1]$ is therefore highest for a candidate that is simultaneously clear of forbidden region and already touching the object. We never use it as a hard constraint, only as a score that ranks and prunes candidates.

\subsubsection{Particle steering (Figure~\ref{fig:architecture}b, \ding{196}).} We sample $N$ trajectories in parallel. A best-of-$N$ baseline selects among them only at the end, spending its budget on trajectories the potential can already condemn long before the last step, since one drifting toward the forbidden region rarely recovers on its own. We therefore let the $N$ trajectories interact as a particle population, resampling at a schedule of steps $\mathcal{R} = \{t_1 < t_2 < \cdots < t_n\}$ ($n=2$ in our implementation). At each $t_i \in \mathcal{R}$ we assign particle $k$ an importance weight
\begin{equation}
\label{eq:weights}
\omega_i^{(k)} \propto \exp\!\big(\lambda\,\Phi^{(k)}_{\max,i}\big), \qquad \Phi^{(k)}_{\max,i} = \max_{j \le i}\,\Phi\big(x_{t_j}^{(k)}\big),
\end{equation}
where $\lambda > 0$ controls how sharply the weights favor high-potential particles. Taking the running maximum lets a particle that looked safe at an earlier checkpoint keep credit for it, rather than be penalized for a single bad look-ahead later. We then resample with replacement in proportion to $\omega_i^{(k)}$, pruning trajectories that the potential predicts will violate the constraint and replicating those it predicts will not. This resampling follows the max-potential design of Feynman-Kac steering~\cite{singhal2025fk}, applied as a corrector~\cite{skreta2025feynmankac} on top of the CFG-guided velocity as the proposal, with the running maximum used as a heuristic score rather than an exact Feynman-Kac potential.

\subsubsection{Churn (Figure~\ref{fig:architecture}b, \ding{197}).} We define \emph{churning} as replacing the deterministic Euler update with a marginal-preserving stochastic one at \emph{every} denoising step rather than only after a resample. Without it, the deterministic ODE feeds a duplicated particle identical inputs at every later step, so resampled twins never diverge. The churned update is
\begin{equation}
\label{eq:churn}
x_{t+dt} = x_t + \big(v_\theta - \eta\,\hat\epsilon_\theta\big)dt + \sqrt{2\eta(1-t)\,dt}\;\epsilon,
\end{equation}
where $\epsilon \sim \mathcal{N}(0,I)$, $dt$ is the step size, and $\hat\epsilon_\theta = x_t - t\,v_\theta$ is the noise endpoint implied by the same rectified path as Eq.~(\ref{eq:lookahead}), here read off over the full state rather than the contact channels alone, with both $v_\theta$ and $\hat\epsilon_\theta$ evaluated at $(x_t, t)$. This is the score-corrected Euler-Maruyama scheme for conditional flow matching~\cite{mark2025feynmankacflow}, with noise scale $\sigma(t) = \sqrt{2\eta(1-t)}$ and a single churn-strength hyperparameter $\eta \ge 0$ that recovers the deterministic ODE exactly at $\eta = 0$. The drift correction $-\eta\,\hat\epsilon_\theta$ compensates the added diffusion, so the marginal $p_t$ each particle targets is unchanged and only the joint trajectory differs, which is what decides whether duplicates stay identical.

\subsubsection{Final selection (Figure~\ref{fig:architecture}b, \ding{198}).} One grasp is still picked from the $N$ survivors. We reconstruct each hand mesh by forward kinematics and keep the candidates that stay at least a fixed margin from the forbidden region $\mathcal{F}$. Among these we return the one closest to the object surface, falling back to the least-violating candidate under a progressively relaxed margin if none qualifies. This exact check is worth its one-time cost because $\Phi$ is a cheap proxy read off contact coordinates rather than the hand mesh, accurate enough to prune particles mid-generation but not to decide the final pass or fail.

\begin{algorithm}[t]
\caption{Inference-time sampling with CFG and SMC.}
\label{alg:steering}
\begin{algorithmic}[1]
\REQUIRE DiT $v_\theta$, prompts $A, B$, forbidden region $\mathcal{F}$ (\S\ref{sec:method-parsing}), CFG $(w, \mathrm{gate}, [\mathrm{lo}, \mathrm{hi}])$, SMC $(\lambda, \mathcal{R})$, churn $\eta$, margin $m$, $S$ steps, $dt{=}1/S$
\STATE $x \sim \mathcal{N}(0,I)$, \quad $\Phi_{\max} \leftarrow 0$
\FOR{$i = 0, \ldots, S-1$}
   \STATE $t \leftarrow i \cdot dt$
   \STATE $v \leftarrow v_\theta(x, t;\, A)$
   \STATE $\hat{x}_c \leftarrow x_c + (1-t)\,v_c$ \COMMENT{Eq.~(\ref{eq:lookahead})}
   \IF{$i \in [\mathrm{lo}, \mathrm{hi}]$}
      \STATE $v^B \leftarrow v_\theta(x, t;\, B)$
      \STATE $g_\perp \leftarrow \mathrm{orth}\big(v_c - v_c^B,\; v_c\big)$
      \STATE $v_c \leftarrow v_c + w\,\mathbf{1}\big[d(\hat{x}_c, \mathcal{F}) < \mathrm{gate}\big]\,g_\perp$ \COMMENT{Eq.~(\ref{eq:guidance})}
   \ENDIF
   \IF{$i \in \mathcal{R}$}
      \STATE $\Phi_{\max} \leftarrow \max\big(\Phi_{\max},\, \Phi(\hat{x}_c)\big)$
      \STATE $x, v \leftarrow \textsc{Resample}\big(x, v;\; \exp(\lambda\,\Phi_{\max})\big)$ \COMMENT{Eq.~(\ref{eq:weights})}
   \ENDIF
   \STATE $x \leftarrow \textsc{Churn}\big(x, v, t, dt, \eta\big)$ \COMMENT{Eq.~(\ref{eq:churn})}
\ENDFOR
\RETURN $\textsc{SelectBest}\big(x;\, \mathcal{F}, \mathcal{O}, m\big)$ \COMMENT{Final selection}
\end{algorithmic}
\end{algorithm}

%% file: Sections/experiments.tex
\section{Experiments}
We first describe NegGrasp, the benchmark we construct for this setting (Section~\ref{exp:dataset}), and the experimental setup (Section~\ref{exp:setup}). We then compare against language-driven grasping baselines (Section~\ref{sec:mainresults}), ablate each component of the inference-time stack (Section~\ref{sec:ablation}), and analyze the steering hyperparameters (Section~\ref{sec:analysis}).

\subsection{NegGrasp dataset}
\label{exp:dataset}
As no existing benchmark evaluates negation-constrained grasping, we augment DexGYSNet~\cite{wei2024dexgys} with prohibitive constraints. For each scene we mine \emph{contested} parts, namely parts that some ground-truth grasps contact while others clearly avoid, so that forbidding such a part rules out natural grasps yet leaves at least one compliant grasp available. Each avoiding grasp is paired with the contested part as its forbidden region, and Gemma 4~\cite{gemma4_2026} converts each pair into a natural-language instruction (e.g., ``grasp the mug, but never touch the handle''). The generation prompts vary tone, clause order, and the choice among twelve avoidance expressions, so no single phrase covers more than a quarter of the split. The benchmark comprises 6{,}928 instructions over 672 scenes and 20 categories, of which 2{,}793 form the test split, inheriting the object-level split of DexGYSNet so that all test instances are unseen during training. Evaluation uses ground-truth part labels from OakInk~\cite{YangCVPR2022OakInk}, while all methods must ground the forbidden part from language alone at inference. Appendix~\ref{app:datasets} details construction and statistics.
 
\subsection{Experimental Setup}
\label{exp:setup}

\subsubsection{Implementation Details.}
Our generator is a 12-layer DiT with hidden dimension 768 and 130M trainable parameters, conditioned on frozen Qwen2.5-0.5B~\cite{qwen2024report} and PartField~\cite{liu2025partfield} encoders. We train it by rectified-flow matching on the positive instructions of DexGYSNet alone, using AdamW with learning rate $2{\times}10^{-4}$, batch size 192, and 1K warmup steps followed by cosine decay. At inference we integrate 16 particles over 25 denoising steps, with guidance weight $w{=}4$ under a 50\,mm contact gate and resampling at steps 10 and 20 with $\lambda{=}20$. The churn strength is $\eta{=}0.1$, and the saturation distances of the potential are $\tau{=}2$\,cm and $\delta{=}3$\,cm. All steering hyperparameters were fixed on a small validation split and held constant across all experiments. Sensitivity analyses for the remaining hyperparameters are in Appendix~\ref{app:hyperparameter}.
 
\subsubsection{Evaluation Metrics.}
Standard grasp metrics are negation-agnostic, since a grasp that firmly seizes the forbidden part still scores well on stability and intention alignment. We therefore evaluate along three complementary axes, using a single 4\,mm threshold for every contact test. \emph{Constraint compliance} is measured by the violation rate (Viol.), the fraction of cases where the hand reaches the forbidden part, by contamination (Cont.), the fraction of hand-object contacts that land on it, and by clearance (Clear.), the median hand-to-forbidden-part distance. \emph{Overall success} is captured by our primary metric, the constraint-aware success rate (CSR), which credits a grasp only when it avoids the forbidden part, touches the instructed part, and contacts the object. \emph{Physical feasibility} is the task success rate (TSR), which credits a grasp only when it is stable in Isaac Gym, following the protocol of DextER~\cite{lee2026dexter}, and free of violation, so that stability bought by seizing the forbidden part earns no credit. We report macro averages over 27 category-part cells to correct for object imbalance, with formal definitions and micro-averaged results in Appendix~\ref{app:metrics}.

\begin{table*}[t]
\centering
\small
\setlength{\tabcolsep}{6pt}
\begin{tabular}{l cc ccc c c}
\toprule
& & & \multicolumn{3}{c}{\textbf{Constraint Compliance}} & \textbf{Overall} & \textbf{Physical} \\
\cmidrule(lr){4-6} \cmidrule(lr){7-7} \cmidrule(lr){8-8}
\textbf{Method} & \textbf{Instruction} & $N$
& Viol.\,$\downarrow$ & Cont.\,$\downarrow$ & Clear.\,$\uparrow$
& CSR\,$\uparrow$ & TSR\,$\uparrow$ \\
& & & (\%) & (\%) & (mm) & (\%) & (\%) \\
\midrule
DexGYS~\cite{wei2024dexgys} & mixed   & 1  & 55.3 & 28.2 & 10.5 & 43.6 & 27.6 \\
DexGYS + BoN                & decomp. & 16 & 36.2 & 14.9 & 14.0 & 56.0 & 36.0 \\
DextER~\cite{lee2026dexter} & mixed   & 1  & 57.9 & 33.4 & 12.5 & 34.1 & 22.4 \\
DextER + BoN                & decomp. & 16 & 29.9 & 14.4 & \textbf{22.3} & 51.5 & 31.0 \\
\textbf{Ours (BoN)}         & decomp. & 16 & \underline{19.6} & \underline{8.5} & 19.8 & \underline{58.6} & \underline{38.3} \\
\textbf{Ours (BoN + Steering)} & decomp. & 16
& \textbf{17.2} & \textbf{6.6} & \underline{21.1} & \textbf{61.5} & \textbf{38.7} \\
\bottomrule
\end{tabular}
\caption{\textbf{Main results on NegGrasp.} Macro-averaged over 27 category--part cells. \emph{Mixed} denotes the raw instruction and \emph{decomp.} the LLM-decomposed input, BoN the scaffolding of Section~\ref{sec:method} over $N$ candidates, and Steering our inference-time guidance combining CFG (Section~\ref{sec:method-cfg}) and SMC (Section~\ref{sec:method-smc}). \textbf{Bold} marks the best and \underline{underline} the second best.}
\label{tab:main}
\end{table*}

\subsection{Main Results}
\label{sec:mainresults}
Table~\ref{tab:main} compares our method against DexGYS and DextER on NegGrasp. Every row marked \emph{decomp.} shares the scaffolding of Section~\ref{sec:method}, where Gemma 4 decomposes the instruction, Find3D~\cite{ma2025find3d} grounds the part to avoid, and the constraint-aware check selects one of 16 candidates. Applying it to each baseline (+ BoN) separates the generator from the scaffolding, and our full method adds steering on top.

\subsubsection{Constraint compliance.}
Both baselines largely ignore the prohibition when given the raw mixed instruction, violating the forbidden part in more than half of all cases (55.3\% and 57.9\%). The scaffolding alone cuts this to 36.2\% and 29.9\%, confirming that decomposition and selection are model agnostic and rescue even negation blind generators. Our full method is the lowest on both violation (17.2\%) and contamination, at 6.6\% against 14.4\% for the strongest baseline configuration, while its clearance of 21.1\,mm is on par with the best at 22.3\,mm. Our generator under the same scaffolding, Ours (BoN), already outperforms every baseline configuration on both, even though it never observes the negation, because the dual conditioning grounds the instructed part in the object geometry so that candidates rarely stray onto the forbidden region.

\subsubsection{Overall success and physical feasibility.}
Our full method achieves the best CSR of 61.5\%, a margin of 5.5 points over the strongest baseline configuration, of which steering contributes 2.9 points over our scaffolded generator alone. Because CSR jointly requires avoidance, instructed part contact, and object contact, these compliance gains are evidently not purchased by abandoning the task. TSR is best as well (38.7\%), with our scaffolded generator second (38.3\%), and the near identical pair shows that steering does not degrade the physical quality of the grasps it steers, a balance we examine in Section~\ref{sec:ablation}.

\subsection{Ablation Study}
\label{sec:ablation}

Table~\ref{tab:ablation} holds the trained model, the particle budget $N{=}16$, and the certifier fixed, so every gap in the table is attributable to the sampling procedure alone.

\begin{table}[t]
\centering
\small
\setlength{\tabcolsep}{4pt}
\resizebox{\columnwidth}{!}{%
\begin{tabular}{l ccc c c}
\toprule
\textbf{Method}
& Viol.\,$\downarrow$ & Cont.\,$\downarrow$ & Clear.\,$\uparrow$
& CSR\,$\uparrow$ & TSR\,$\uparrow$ \\
\midrule
BoN                      & 19.6 & 8.5  & 19.8 & 58.6 & 38.3 \\
+\,CFG                   & \underline{17.8} & \underline{6.7} & \textbf{22.2} & 61.0 & 31.0 \\
+\,SMC (deterministic)   & 28.3 & 12.9 & 20.0 & \textbf{61.8} & \textbf{40.5} \\
+\,Churn (\textbf{Ours}) & \textbf{17.2} & \textbf{6.6} & \underline{21.1} & \underline{61.5} & \underline{38.7} \\
\bottomrule
\end{tabular}}
\caption{\textbf{Ablation of the inference-time stack.} Each row adds one component to the row above, from best-of-16 selection with no intervention. Macro averages on NegGrasp.}
\label{tab:ablation}
\end{table}

\subsubsection{CFG buys avoidance and pays in physics.}
Negative prompt CFG shifts the proposal distribution away from the forbidden region and improves every constraint metric, most visibly widening the median clearance from 19.8\,mm to 22.2\,mm. The cost appears on the physical axis, where TSR drops from 38.3 to 31.0, since guidance perturbs the velocity field away from the learned distribution and the resulting particles avoid the forbidden part but grip the object less firmly.

\subsubsection{Deterministic SMC trades particle collapse for physics.}
Adding SMC resampling recovers the physical axis, raising TSR to 40.5, the best value in the table. As described in Section~\ref{sec:method-smc}, however, clones duplicated under the deterministic ODE never diverge again, and the particle set collapses from 16 to between 2 and 5 unique samples, the only row in the table that loses diversity. The few survivors are precisely those favored by the object proximity term of the reward, so violations flow back in and the violation rate jumps to 28.3, worse than the no intervention baseline.

\subsubsection{Churn blocks the collapse and keeps both gains.}
Replacing the ODE step with the marginal preserving SDE of Section~\ref{sec:method-smc} separates clones within a single step without altering the distribution that each particle samples from. All 16 particles remain unique through the entire trajectory, and the full method attains the avoidance of CFG together with the constrained success of SMC. The components are complementary, since CFG alone sacrifices physical quality, SMC alone reintroduces violations through collapse, and churn alone has nothing to steer.

\subsection{Analysis}
\label{sec:analysis}
 
\subsubsection{Resampling schedule.}
We treat the number of resampling events $n=|\mathcal{R}|$ as a free hyperparameter, placing multiple events at uniform intervals over the 25 denoising steps. Table~\ref{tab:sweep} sweeps $n$ from 0 to 4 together with the placement of single events. Churn keeps all 16 particles unique throughout, so the schedule can be varied freely without risking collapse.

The constraint metrics respond only mildly, with violation staying near the CFG level up to $n{=}2$ and drifting upward once events accumulate, since every event applies the object proximity reward once more. The separation that matters appears on TSR. A single event trades the two success metrics against each other depending on placement, as the early $\{5\}$ reaches a TSR of 36.2 at the lowest CSR of the sweep while the late $\{20\}$ reaches a strong CSR of 63.1 at the worst TSR of 28.5. Our schedule at $n{=}2$ is the only configuration near the top of both columns at once, and it leads TSR by a clear margin. Additional events do not pay, since the extra violations at $n{=}3$ and $n{=}4$ cancel their physical gains.

\subsubsection{CFG strength.}
Table~\ref{tab:sweep} sweeps the guidance weight with every other setting frozen, and the two axes peak at different points. Compliance is already strong at $w{=}4$, which attains the lowest violation rate, and pushing further buys little, since clearance and CSR keep rising to $w{=}16$ while violation drifts back up. The physical axis instead reverses at once, with TSR peaking at $w{=}4$ and falling to 34.8 and 31.1 at $w{=}6$ and $w{=}16$ as stronger guidance pushes samples further from the learned distribution, the same trade observed in Section~\ref{sec:ablation}. We therefore adopt $w{=}4$, beyond which the physical axis degrades faster than compliance improves, and we attribute the robustness even at $w{=}16$ to the contact gate, which fires only near the forbidden region.

\begin{table}[t]
\centering
\small
\setlength{\tabcolsep}{4pt}
\resizebox{\columnwidth}{!}{%
\begin{tabular}{l ccc c c}
\toprule
\textbf{Setting}
& Viol.\,$\downarrow$ & Cont.\,$\downarrow$ & Clear.\,$\uparrow$
& CSR\,$\uparrow$ & TSR\,$\uparrow$ \\
\midrule
\rowcolor{gray!15}
\multicolumn{6}{l}{\emph{Schedule} $\mathcal{R}$ (at $w{=}4$)} \\
\quad $\emptyset$ ($n{=}0$)                & 17.8 & \underline{6.7} & \textbf{22.2} & 61.0 & 31.0 \\
\quad $\{5\}$ ($n{=}1$, early)             & 17.9 & 7.3 & 20.4 & 59.5 & \underline{36.2} \\
\quad $\{20\}$ ($n{=}1$, late)             & \textbf{16.9} & \textbf{6.6} & \underline{21.2} & \underline{63.1} & 28.5 \\
\quad $\{10,20\}$ ($n{=}2$, \textbf{Ours}) & \underline{17.2} & \textbf{6.6} & 21.1 & 61.5 & \textbf{38.7} \\
\quad $\{6,12,18\}$ ($n{=}3$)              & 19.7 & 8.5 & 19.5 & \textbf{63.5} & 31.6 \\
\quad $\{5,10,15,20\}$ ($n{=}4$)           & 19.2 & 8.0 & \underline{21.2} & 61.6 & 33.3 \\
\midrule
\rowcolor{gray!15}
\multicolumn{6}{l}{\emph{Guidance} $w$ (at $\mathcal{R}{=}\{10,20\}$)} \\
\quad $w{=}2$                 & 20.0 & \underline{6.5} & 18.7 & 57.2 & 31.3 \\
\quad $w{=}4$ (\textbf{Ours}) & \textbf{17.2} & 6.6 & \underline{21.1} & 61.5 & \textbf{38.7} \\
\quad $w{=}6$                 & \underline{18.4} & 7.1 & \underline{21.1} & \underline{62.8} & \underline{34.8} \\
\quad $w{=}16$                & 18.7 & \textbf{5.7} & \textbf{23.7} & \textbf{65.6} & 31.1 \\
\bottomrule
\end{tabular}}
\caption{\textbf{Steering hyperparameters.} Each block sweeps one axis with every other setting frozen, and best and second best are marked within each block. Macro averages on NegGrasp.}
\label{tab:sweep}
\end{table}
 

%% file: Sections/conclusion.tex
\section{Conclusion}
We showed that language-driven dexterous grasping fails systematically once an instruction says what not to touch, and addressed this entirely at inference time, steering a flow-matching grasp DiT with contact-gated CFG and churn-diversified SMC resampling, without a single negative training example. On NegGrasp, our benchmark for this setting, this cuts the violation rate of the strongest baseline to less than a third while improving both constraint-aware and physical success.

%% file: Sections/appendix_notation.tex
\section{Notation}
\label{app:notation}
Table~\ref{tab:notation} summarizes the notation used in Section~\ref{sec:method} and Algorithm~\ref{alg:steering}.

\begin{table}[t]
\centering
\small
\begin{tabular}{ll}
\toprule
Symbol & Meaning \\
\midrule
\multicolumn{2}{l}{\textit{State and flow}} \\
$x = (x_g, x_c)$ & generation state, Eq.~(\ref{eq:state}) \\
$x_g \in \mathbb{R}^{28}$ & grasp pose, a 3D translation, a 3D axis-angle \\
 & rotation, and 22 ShadowHand joint angles \\
$x_c \in \mathbb{R}^{8 \times 4}$ & contact tokens, 8 slots of a 3D coordinate \\
 & and an occupancy flag \\
$t, dt, S$ & flow time, step size $dt = 1/S$, number of \\
 & denoising steps \\
$x_0, x_1$ & noise endpoint $x_0 \sim \mathcal{N}(0,I)$ and data \\
 & endpoint of the rectified path \\
$v_\theta(x,t)$ & velocity field regressed by the DiT \\
$\hat{x}_c$ & look-ahead contact estimate, Eq.~(\ref{eq:lookahead}) \\
$\hat\epsilon_\theta$ & noise-endpoint estimate, $x_t - t\,v_\theta$ \\
\midrule
\multicolumn{2}{l}{\textit{Grounding and potential}} \\
$\mathcal{F}, \mathcal{O}$ & forbidden region and object surface \\
$d(\cdot,\cdot)$ & minimum Euclidean distance \\
$\psi$ & normalized proximity score, Eq.~(\ref{eq:potential}) \\
$\Phi \in [0,1]$ & look-ahead potential, Eq.~(\ref{eq:potential}) \\
$\tau, \delta$ & saturation distances of $\psi$ for $\mathcal{F}$ and $\mathcal{O}$ \\
\midrule
\multicolumn{2}{l}{\textit{Guidance}} \\
$A, B$ & positive and negative prompts \\
$v_c^A, v_c^B$ & contact velocities predicted under $A$ and $B$ \\
$w, \mathrm{gate}$ & guidance weight and distance gate \\
$[\mathrm{lo}, \mathrm{hi}]$ & steps over which guidance is applied \\
$\mathrm{orth}(g,a)$ & component of $g$ orthogonal to $a$ \\
$g_\perp$ & orthogonal guidance term, Eq.~(\ref{eq:guidance}) \\
\midrule
\multicolumn{2}{l}{\textit{Particle steering and churn}} \\
$N, k$ & number of particles and particle index \\
$\omega^{(k)}, \lambda$ & importance weight, Eq.~(\ref{eq:weights}), and \\
 & weight sharpness \\
$\mathcal{R}$ & resampling schedule $\{t_1, \ldots, t_n\}$ \\
$\Phi^{(k)}_{\max}$ & running maximum of $\Phi$ for particle $k$ \\
$\eta, \sigma(t)$ & churn strength and noise scale \\
 & $\sigma(t) = \sqrt{2\eta(1-t)}$, Eq.~(\ref{eq:churn}) \\
$m$ & clearance margin of the final selection \\
\bottomrule
\end{tabular}
\caption{Notation used in Section~\ref{sec:method}.}
\label{tab:notation}
\end{table}

\section{Metrics}
\label{app:metrics}

This appendix formalizes the metrics of Section~\ref{exp:setup} and specifies how scores are aggregated across the imbalanced test split.

\subsection{Formal Definitions}
\label{app:metrics-defs}
For test case $i$, let $H_i$ denote the surface points of the predicted hand mesh, posed by forward kinematics from the predicted grasp. Let $O_i$ be the object point cloud, $\mathcal{F}_i \subset O_i$ the ground-truth forbidden part, and $d(A,B)$ the minimum pairwise Euclidean distance. All contact tests share a single threshold $\tau_v$, and we use $\tau_v = 4$\,mm throughout.

\paragraph{Violation rate.} A case counts as a violation if any hand point reaches the forbidden part:
\begin{equation}
\mathrm{Viol}_i = \mathbf{1}\big[\, d(H_i, \mathcal{F}_i) < \tau_v \,\big]. \tag{7}
\end{equation}

\paragraph{Contamination.} Let $C_i = \{\, p \in O_i : d(H_i, \{p\}) < \tau_v \,\}$ be the set of object points the hand contacts. Contamination is the fraction of those contacts that land on the forbidden part,
\begin{equation}
\mathrm{Cont}_i = \frac{|\, C_i \cap \mathcal{F}_i \,|}{|C_i|}, \tag{8}
\end{equation}
defined as $0$ when $C_i$ is empty.

\paragraph{Clearance.} $\mathrm{Clear}_i = d(H_i, \mathcal{F}_i)$, the safety margin of the hand from the forbidden part. It is reported as a median rather than a mean, since a single distant hand would otherwise dominate.

\paragraph{Constraint-aware success rate.} Let $T_i$ be the target set: the parts that the ground-truth grasp of case $i$ contacts within $\tau_v$, minus the forbidden part. With $\mathrm{Hit}_i = \mathbf{1}[\text{the prediction contacts some part in } T_i \text{ within } \tau_v]$ and $\mathrm{Contact}_i = \mathbf{1}[d(H_i, O_i) < \tau_v]$,
\begin{equation}
\mathrm{CSR}_i = (1 - \mathrm{Viol}_i)\cdot \mathrm{Hit}_i \cdot \mathrm{Contact}_i . \tag{9}
\end{equation}
$T_i$ realizes the instructed part geometrically: each instruction is generated from a reference compliant grasp (Appendix~\ref{app:datasets-construction}), so the parts that grasp contacts are the parts the instruction directs the hand toward.

\paragraph{Task success rate.} $\mathrm{TSR}_i = \mathrm{Stable}_i \cdot (1-\mathrm{Viol}_i)$, where $\mathrm{Stable}_i$ is the Isaac Gym stability outcome under the simulation protocol of DextER~\cite{lee2026dexter}.

\subsection{Aggregation: Macro and Micro}
\label{app:metrics-agg}
Evaluation covers the 2{,}793 of 2{,}843 test cases for which Find3D returns a grounding, and this case set $S$ is shared by every method. A cell is a (category, forbidden part) pair; let $\mathcal{C}$ be the set of the 27 cells the evaluated cases span, and $S_c \subseteq S$ the cases falling in cell $c$. For any per-case score $m_i$ (one of $\mathrm{Viol}_i$, $\mathrm{Cont}_i$, $\mathrm{CSR}_i$, $\mathrm{TSR}_i$), macro first averages within each cell and then averages the 27 resulting cell scores, while micro averages $m_i$ directly over all evaluated cases:
\begin{align}
m^{\text{macro}} &= \frac{1}{|\mathcal{C}|}\sum_{c \in \mathcal{C}} \frac{1}{|S_c|}\sum_{i \in S_c} m_i, \tag{10}\label{eq:macro}\\
m^{\text{micro}} &= \frac{1}{|S|}\sum_{i \in S} m_i. \tag{11}\label{eq:micro}
\end{align}
Clearance replaces the innermost case-average with a median (over $S_c$ in macro's inner sum, over $S$ for micro), but macro's outer average over cells is still a mean even for Clearance. The distinction matters because the test split is heavily imbalanced: the mug category alone contributes 1{,}794 cases (64\% of the evaluated cases), and since every mug case shares the same forbidden part (mug\_handle), all of them fall into a single one of the 27 cells. Micro scores therefore largely reflect one category, and the main paper reports macro scores. Unless explicitly marked otherwise, every number in this paper is a macro score. Micro scores appear only in this appendix section and are labeled as such.

\subsection{Micro-Averaged Results}
\label{app:metrics-micro}
Table~\ref{tab:micro} reports micro scores for every row of Table~\ref{tab:main}. The ranking of the two Ours variants is unchanged under micro averaging, and the margins over the baselines widen because mug cases dominate the pooled set. Relative to the strongest baseline configuration, DextER + BoN, the full method reduces violation by 84\% under micro averaging (43.0\% to 6.9\%) versus 42\% under macro averaging (29.9\% to 17.2\%).

\begin{table}[t]
\centering
\small
\setlength{\tabcolsep}{4pt}
\resizebox{\columnwidth}{!}{%
\begin{tabular}{l cc cc c}
\toprule
\textbf{Method} & Viol.\,$\downarrow$ & Cont.\,$\downarrow$ & Clear.\,$\uparrow$ & CSR\,$\uparrow$ & TSR\,$\uparrow$ \\
& (\%) & (\%) & (mm) & (\%) & (\%) \\
\midrule
DexGYS (mixed)              & 56.1 & 18.3 & 2.3  & 43.1 & 28.8 \\
DexGYS + BoN                & 26.0 & 7.5  & 8.8  & \underline{68.1} & \textbf{47.5} \\
DextER (mixed)              & 70.5 & 43.0 & 0.3  & 25.0 & 24.1 \\
DextER + BoN                & 43.0 & 27.8 & 4.6  & 30.2 & 35.5 \\
Ours (BoN)                  & \underline{7.7}  & \underline{2.8}  & \underline{19.2} & 66.6 & 41.5 \\
Ours (BoN + Steering)       & \textbf{6.9}  & \textbf{2.3}  & \textbf{21.9} & \textbf{70.2} & \underline{41.6} \\
\bottomrule
\end{tabular}}
\caption{\textbf{Micro-averaged results.} All numbers are computed over the same 2{,}793 evaluated cases as the main tables (pooled means, with the pooled median for Clearance). \textbf{Bold} marks the best and \underline{underline} the second best.}
\label{tab:micro}
\end{table}

\section{Datasets}
\label{app:datasets}

This appendix quantifies the absence of negation in the training corpus and details the construction of NegGrasp.

\subsection{Negation is Essentially Absent from DexGYSNet}
\label{app:datasets-scarcity}
We scanned all 49{,}433 instructions of DexGYSNet (39{,}462 train, 9{,}971 test) for negation cues with the case-insensitive word list \{\emph{avoid, not, never, without, do not, don't, except, refrain, away from}\}. Only 23 instructions (0.047\%) contain any cue. Manual inspection of all 23 shows that 18 negate an \emph{outcome} rather than a contact region (``avoid dropping it,'' ``without spilling,'' ``to avoid injury''), and only 5 (0.010\%) forbid touching a named object part (e.g., ``gripping its handle with 5 fingers to avoid the blade''). At one part-directed avoidance instruction per ten thousand, the corpus is far too sparse to induce the association between negation words and avoidance behavior, which motivates both our benchmark and our inference-time treatment of negation.


\subsection{NegGrasp Construction}
\label{app:datasets-construction}
NegGrasp augments DexGYSNet with prohibitive constraints. Each benchmark case consists of a DexGYSNet scene, one object part that must not be touched, and a newly written instruction that asks for a grasp while forbidding that part. Only the language is new: every case is built around a ground-truth grasp that already satisfies its prohibition, so failure on the benchmark reflects a failure to honor the instruction rather than an inability to produce the required grasp. Construction proceeds in four steps.
\begin{enumerate}
\item \textbf{Contested-part mining.} DexGYSNet provides many ground-truth grasps per scene, and different grasps hold the same object by different parts, which is what makes a meaningful prohibition possible. Forbidding a part that no grasp touches (the bottom of a mug) is vacuous, since a model that ignores the instruction still complies. Forbidding a part that every grasp touches (the body of a bowl) leaves the case unsolvable. We therefore forbid only \emph{contested} parts: parts that at least one grasp touches (hand-to-part distance below 4\,mm) and at least one other grasp clearly avoids (above 10\,mm), where the gap between the two thresholds keeps borderline grasps out of both groups. Forbidding a contested part rules out grasps that naturally occur, while a compliant grasp is proven to exist, with a comfortable margin (median 26.4\,mm).
\item \textbf{Pairing.} Each avoiding grasp is paired with the contested part it avoids. The pair fixes everything about a case except its instruction: the part becomes the forbidden region, and the grasp serves as the reference compliant grasp, an existence proof that the instruction written in the next step can be satisfied.
\item \textbf{Instruction generation.} Gemma 4~\cite{gemma4_2026} writes one command sentence per case. It is not asked to invent any fact about the grasp: its input is the reference grasp's measured contact record, namely which finger links touch which part, the name of the forbidden part, and the clearance the grasp keeps from it, and its only task is to phrase these facts as a natural command. Variety comes from randomized writing directives covering five tones, three clause orders, and twelve avoidance expressions, with a brief reason for the prohibition (``it could spill'') requested in some cases. Every instruction forbids the part. Whether it also names where to grasp is left to the model, and roughly three quarters of the sentences do. The complete system prompt, the per-case user-message template, and its randomization directives are shown in Figure~\ref{fig:instr-gen-prompt}.
\item \textbf{Filtering and deduplication.} A generated sentence is discarded unless it names the forbidden part with an explicit negative expression. To keep sentences from echoing one another, each new sentence is compared against the accepted sentences of its (category, part) group by 4-gram overlap and regenerated when the similarity exceeds 0.55. The released split contains no two identical sentences (median within-group similarity 0.292). Instructions average 107 characters, and 15\% state a reason for the prohibition.
\end{enumerate}

\textbf{Statistics.} The benchmark comprises 6{,}928 instructions (4{,}085 train, 2{,}843 test) over 672 scenes, 787 unique (scene, part) pairs, and 20 object categories, inheriting the object-level split of DexGYSNet so that every test object instance is unseen during training. The test split spans 135 objects, 15 categories, and 27 (category, forbidden part) cells. The train split is not used by any experiment in this paper: our generator is trained on DexGYSNet's original positive instructions (Section~\ref{sec:method-arch}), no baseline is fine-tuned on NegGrasp, and every number we report is computed on the test split alone.

\section{LLM Instruction Parser}
\label{app:parser}
The parser receives the object category and the raw instruction as the user message
\begin{quote}
\ttfamily Object category: \{category\}\\
Instruction: "\{instruction\}"
\end{quote}
and must return only a JSON object with two fields: \texttt{positive\_instruction} (the rewrite with any avoidance clause stripped) and \texttt{parts}, a list of \{\texttt{name}, \texttt{forbidden}\} entries covering the complete part vocabulary of the object. Outputs are decoded under a constrained JSON schema and validated for lowercase snake\_case names without repeated words. On validation failure, the query is reissued with the error message appended, for up to three attempts at decreasing temperature. On the test split, the parser flags more than one part as forbidden in 80 of 2{,}843 cases, where the first flagged part is used, and flags none in 6. The latter fall among the 50 cases excluded from evaluation because no forbidden part can be grounded (Appendix~\ref{app:metrics-agg}).

The system prompt specifies three groups of rules. \emph{Naming}: each part is a snake\_case label of the form \texttt{\{object\_noun\}\_\{part\_name\}} with a shortened object noun and no duplicated words, and a genuinely bilateral component is split into \texttt{left\_}/\texttt{right\_} entries, since an avoidance clause may target one side. \emph{Granularity}: the parser is instructed to prefer fewer, larger parts. \emph{Forbidden flagging}: only a part introduced by an explicit negative word is flagged, and a part named as where to grasp is never flagged. A representative few-shot example:

\begin{lstlisting}[
  numbers=none,
  breaklines=true,
  basicstyle=\scriptsize\ttfamily,
  frame=single,
  framesep=6pt,          % 프레임과 코드 사이 안쪽 여백
  xleftmargin=4pt,
  xrightmargin=4pt,
  aboveskip=1.2\baselineskip,   % 본문과의 위쪽 간격
  belowskip=1.2\baselineskip,   % 본문과의 아래쪽 간격
  backgroundcolor=\color{gray!5} % 아주 옅은 배경 (선택)
]
category = "teapot",
instruction = "Grasp the teapot by the handle
               but avoid the spout."
->
{
  "positive_instruction":
      "Grasp the teapot by the handle.",
  "parts": [
    {"name": "teapot_handle", "forbidden": false},
    {"name": "teapot_spout",  "forbidden": true },
    {"name": "teapot_body",   "forbidden": false},
    {"name": "teapot_lid",    "forbidden": false}
  ]
}
\end{lstlisting}

The complete system prompt is shown in Figure~\ref{fig:parser-prompt} and included in the released inference code. We note that the parser, like the rest of the decomposition-and-grounding scaffolding, is shared by every compared method: any effect of prompt design on grounding quality applies to all rows marked decomp.\ in Table~\ref{tab:main} equally, and does not enter the comparison between methods.

\input{Sections/appendix_hyperparameter}
\input{Sections/appendix_model}
\input{Sections/appendix_qulitative}

\begin{figure*}[p]
\begin{lstlisting}[
  numbers=none,
  breaklines=true,
  breakindent=0pt,
  basicstyle=\scriptsize\ttfamily,
  columns=fullflexible,
  frame=single,
  framesep=8pt,                  % 프레임-텍스트 안쪽 여백 (겹침 해소)
  xleftmargin=8pt,
  xrightmargin=8pt,
  aboveskip=\baselineskip,
  belowskip=0.5\baselineskip,    % 캡션과의 간격
  backgroundcolor=\color{gray!4},% 옅은 배경 (선택)
  escapechar=@
]
@\textbf{[System Prompt]}@
You are the instruction parser for a language-driven dexterous-grasp robot system. You are given the object's category and one natural-language grasp instruction that may combine a positive target ("grasp the X") with a negative, avoidance clause ("but avoid the Y", "without touching Z", "never the W").

Produce exactly two things:
1. "positive_instruction": the instruction rewritten to describe ONLY where/how to grasp (strip out any negative/avoidance clause; keep it a natural imperative sentence).
2. "parts": the COMPLETE list of this object's real, physically distinct named parts -- every major part a real object of this category has, not just the ones mentioned in the instruction. GRANULARITY (important): most objects have only 1-2 such parts -- a plain object with no separate lid/handle/cap/blade is a SINGLE part (the whole body, e.g. a bowl or cup is just one part -- do not invent a split where the object is one continuous piece); a typical object with one distinguishing feature (a handle, a cap, a blade, a nozzle) is 2 parts; only a genuinely multi-component object (e.g. a teapot with body+spout+handle+lid, or a power drill with chuck+motor+grip+trigger+battery) has 3 or more. Do not pad the list with invented sub-features (screws, pads, hinges, seams, small buttons) that are not prominent, separately load-bearing regions -- when in doubt, prefer FEWER, LARGER parts over more, smaller ones. If (and only if) the object has a genuine LEFT/RIGHT symmetric pair of the same functional component (e.g. two temple arms on eyeglasses, two ear cups on headphones), give each side its own entry prefixed "left_"/ "right_" -- an avoidance instruction may target only one specific side, so do not collapse a symmetric pair into one generic entry. Each entry has:
   - "name": a lowercase snake_case string in the form "{object_noun}_{part_name}", where object_noun is the object's natural singular noun -- often a SHORTENED form of the category, not the literal category string (e.g. category="hair_dryer" -> noun "dryer"; category="cylinder_bottle" -> noun "bottle"). NEVER duplicate a word (e.g. "hair_dryer_dryer_nozzle" is WRONG, it repeats "dryer"; the correct form is "dryer_nozzle"). If the object noun already starts the part's natural name (e.g. category="lotion_pump", part="pump head"), do NOT prefix it again -- just use "pump_head", not "pump_pump_head". A bare single word with no prefix is fine when the word is unambiguous on its own (e.g. "switch", "zipper", "trigger") -- don't force a prefix that makes the name awkward.
   - "forbidden": true if the instruction has an explicit negative/avoidance word ("avoid", "never", "not", "without touching", "but not") targeting this exact part, false otherwise. Exactly one canonical entry per real part -- never list the same physical part twice under two different name spellings.

This part list is used to let a 3D part-grounding model tell the object's parts apart from each other, so it must be the object's actual anatomy (a closed, mutually-exclusive set), not a single vague catch-all -- but also not an over-segmented parts-catalog. These examples (for objects unrelated to whatever you're being asked about) illustrate the expected granularity and NAMING PATTERN, not a lookup table -- work out the real parts and their names for whatever category you're actually given, using this same style:
- category=bowl: ['bowl_body']
- category=backpack: ['backpack_body', 'shoulder_strap', 'zipper']
- category=teapot: ['teapot_spout', 'teapot_handle', 'teapot_lid', 'teapot_body']
- category=headphones: ['headband', 'left_ear_cap', 'right_ear_cap']

CRITICAL: a part mentioned as WHERE TO GRASP is a positive target ("forbidden": false), not a forbidden part. Only a part introduced by an explicit negative/avoidance word belongs with "forbidden": true. If you see no such negative word anywhere in the instruction, every entry's "forbidden" must be false (but the full part list is still filled in).

Examples (again, illustrating the pattern on unrelated objects -- reason about the real parts of whatever object you're actually given):
- category=teapot, instruction="Lift the teapot from its body." -> positive_instruction="Lift the teapot from its body.", parts=[{"name":"teapot_body","forbidden":false}, {"name":"teapot_spout","forbidden":false}, {"name":"teapot_handle","forbidden":false}, {"name":"teapot_lid","forbidden":false}] (the body is where to grasp, not avoid, but the spout/handle/lid are still part of the object's anatomy and belong in the list)
- category=teapot, instruction="Grasp the teapot by the handle but avoid the spout." -> positive_instruction="Grasp the teapot by the handle.", parts=[{"name":"teapot_handle","forbidden":false}, {"name":"teapot_spout","forbidden":true}, {"name":"teapot_body","forbidden":false}, {"name":"teapot_lid","forbidden":false}]
- category=backpack, instruction="Pick up the backpack by the strap, never the zipper." -> positive_instruction="Pick up the backpack by the strap.", parts=[{"name":"shoulder_strap","forbidden":false}, {"name":"zipper","forbidden":true}, {"name":"backpack_body","forbidden":false}]
- category=bowl, instruction="Pick up the bowl." -> positive_instruction="Pick up the bowl.", parts=[{"name":"bowl_body","forbidden":false}] (a plain bowl is one continuous piece -- do NOT invent a rim/base/interior split)

Output ONLY the JSON object matching the required schema. No prose, no markdown fences.

@\textbf{[User Prompt]}@
Object category: {category}
Instruction: "{instruction}"
\end{lstlisting}
\caption{\textbf{System prompt of the LLM instruction parser.} The figure shows the complete system prompt and the per-case user-message template (Appendix~\ref{app:parser}), both reproduced verbatim from the released inference code.}
\label{fig:parser-prompt}
\end{figure*}

\begin{figure*}[t]
\begin{lstlisting}[
  numbers=none,
  breaklines=true,
  breakindent=0pt,
  basicstyle=\scriptsize\ttfamily,
  columns=fullflexible,
  frame=single,
  framesep=8pt,
  xleftmargin=8pt,
  xrightmargin=8pt,
  belowskip=0.5\baselineskip,
  backgroundcolor=\color{gray!4},
  escapechar=@
]
@\textbf{[System Prompt]}@
You write one-sentence natural-language commands for a robot with a five-fingered dexterous hand. Output ONLY the command sentence, no quotes, no explanations.

@\textbf{[User Prompt]}@
Object: {e['class_name']}.
A reference command for this grasp was: "{e['base_guidance']}".
The grasp actually used: {contacts}.
The region that must NOT be touched: the {part} (stays {e['avoid_dist_mm']}mm away).

Write ONE new command telling the robot to grasp the object while never touching the {part}.
Style: {reg}. Sentence order: {order}.
Prefer wording like one of: {vocab} (pick what fits, adapt freely).
[if want_reason] Include a brief reason, e.g. because {reason}.
[if base_guidance mentions part] NOTE: the reference command mentions the {part}, but this grasp does NOT touch it -- trust the actual contacts, not the reference.
Do not copy the reference command; rephrase the intent in your own words.
\end{lstlisting}
\caption{\textbf{Instruction-generation prompt for NegGrasp construction.} The figure shows the complete prompt, reproduced verbatim from the dataset-construction code's \texttt{build\_prompt()} function. The placeholders \texttt{reg}, \texttt{order}, \texttt{vocab}, and \texttt{contacts} are values already resolved before insertion; Appendix~\ref{app:datasets-construction} describes how they are sampled.}
\label{fig:instr-gen-prompt}
\end{figure*}

%% file: Sections/appendix_hyperparameter.tex
\section{Hyperparameter Sensitivity Analysis}
\label{app:hyperparameter}

All steering hyperparameters were fixed once on a small validation split and held constant across every experiment in the main paper. This appendix probes the remaining axes of the inference-time stack around the final configuration ($N{=}16$, $\lambda{=}20$, $\eta{=}0.1$, a 50\,mm contact gate, guidance applied over the full step window $[0, 24]$, and potential saturation distances $\tau{=}20$\,mm and $\delta{=}30$\,mm), varying one axis at a time on the full NegGrasp test split with every other setting frozen. The resampling schedule $\mathcal{R}$ and the guidance weight $w$ are analyzed in the main paper (Table~\ref{tab:sweep}) and are not repeated here. Table~\ref{tab:hparam_sweep} reports macro averages over the 27 category-part cells, computed with the identical evaluation code and case set as the main tables.

\begin{table}[t]
\centering
\small
\setlength{\tabcolsep}{4pt}
\resizebox{\columnwidth}{!}{%
\begin{tabular}{l ccc c c}
\toprule
& \multicolumn{3}{c}{\textbf{Constraint Compliance}} & \textbf{Overall} & \textbf{Physical} \\
\cmidrule(lr){2-4} \cmidrule(lr){5-5} \cmidrule(lr){6-6}
\textbf{Setting}
& Viol.\,$\downarrow$ & Cont.\,$\downarrow$ & Clear.\,$\uparrow$
& CSR\,$\uparrow$ & TSR\,$\uparrow$ \\
& (\%) & (\%) & (mm) & (\%) & (\%) \\
\midrule
\rowcolor{gray!15}
\multicolumn{6}{l}{\emph{Particles} $N$} \\
\quad $N{=}2$                  & 32.8 & 15.5 & 17.5 & 56.3 & 29.9 \\
\quad $N{=}4$                  & 28.8 & 12.5 & 16.6 & 59.3 & \underline{31.6} \\
\quad $N{=}8$                  & 23.2 & 10.6 & 19.6 & \textbf{61.8} & 31.2 \\
\quad $N{=}16$ (\textbf{Ours}) & \textbf{17.2} & \textbf{6.6} & \underline{21.1} & \underline{61.5} & \textbf{38.7} \\
\quad $N{=}32$                 & \underline{17.9} & \underline{8.2} & \textbf{23.0} & 59.6 & 29.2 \\
\midrule
\rowcolor{gray!15}
\multicolumn{6}{l}{\emph{Weight sharpness} $\lambda$} \\
\quad $\lambda{=}5$                  & \textbf{15.8} & 7.3 & \textbf{21.9} & 60.3 & \underline{31.7} \\
\quad $\lambda{=}10$                 & 18.3 & \underline{6.8} & \textbf{21.9} & \underline{61.5} & 26.9 \\
\quad $\lambda{=}20$ (\textbf{Ours}) & \underline{17.2} & \textbf{6.6} & \underline{21.1} & \underline{61.5} & \textbf{38.7} \\
\quad $\lambda{=}50$                 & 19.5 & 9.1 & 20.2 & \textbf{62.9} & 29.9 \\
\midrule
\rowcolor{gray!15}
\multicolumn{6}{l}{\emph{Churn strength} $\eta$} \\
\quad $\eta{=}0$ (deterministic ODE) & 28.3 & 12.9 & 20.0 & 61.8 & \textbf{40.5} \\
\quad $\eta{=}0.05$                  & 19.3 & 7.8 & 21.7 & \textbf{65.9} & 37.5 \\
\quad $\eta{=}0.1$ (\textbf{Ours})   & \underline{17.2} & \underline{6.6} & 21.1 & 61.5 & \underline{38.7} \\
\quad $\eta{=}0.2$                   & 17.3 & \textbf{6.0} & \underline{23.6} & \underline{62.6} & 29.2 \\
\quad $\eta{=}0.3$                   & \textbf{17.1} & 7.1 & \textbf{23.7} & 62.5 & 27.0 \\
\midrule
\rowcolor{gray!15}
\multicolumn{6}{l}{\emph{Contact gate} (mm)} \\
\quad gate\,$=20$                  & 19.7 & 7.5 & 19.6 & 60.2 & \underline{35.4} \\
\quad gate\,$=50$ (\textbf{Ours})  & \textbf{17.2} & \textbf{6.6} & \underline{21.1} & \textbf{61.5} & \textbf{38.7} \\
\quad gate\,$=100$                 & \underline{18.7} & \underline{7.2} & \textbf{22.0} & \underline{60.7} & 29.6 \\
\midrule
\rowcolor{gray!15}
\multicolumn{6}{l}{\emph{Guidance window} $[\mathrm{lo}, \mathrm{hi}]$} \\
\quad $[0, 12]$ (early only)       & 18.9 & 8.1 & 20.6 & \textbf{61.6} & 28.9 \\
\quad $[12, 24]$ (late only)       & \underline{17.7} & \textbf{6.2} & \underline{20.7} & 60.9 & \underline{30.7} \\
\quad $[0, 24]$ (\textbf{Ours})    & \textbf{17.2} & \underline{6.6} & \textbf{21.1} & \underline{61.5} & \textbf{38.7} \\
\midrule
\rowcolor{gray!15}
\multicolumn{6}{l}{\emph{Forbidden saturation} $\tau$ (mm)} \\
\quad $\tau{=}8$                   & 21.8 & \underline{9.2} & 20.6 & 59.3 & 31.7 \\
\quad $\tau{=}20$ (\textbf{Ours})  & \textbf{17.2} & \textbf{6.6} & \underline{21.1} & \textbf{61.5} & \textbf{38.7} \\
\quad $\tau{=}40$                  & 20.1 & 9.7 & \textbf{22.4} & 59.4 & 29.7 \\
\quad $\tau{=}80$                  & \underline{19.2} & 9.6 & 20.0 & \underline{59.9} & \underline{37.1} \\
\midrule
\rowcolor{gray!15}
\multicolumn{6}{l}{\emph{Object saturation} $\delta$ (mm)} \\
\quad $\delta{=}15$                & \underline{20.8} & \underline{8.5} & \textbf{22.3} & \underline{61.0} & 26.6 \\
\quad $\delta{=}30$ (\textbf{Ours}) & \textbf{17.2} & \textbf{6.6} & \underline{21.1} & \textbf{61.5} & \textbf{38.7} \\
\quad $\delta{=}60$                & 22.7 & 8.9 & 21.0 & 59.9 & \underline{29.1} \\
\bottomrule
\end{tabular}}
\caption{\textbf{Hyperparameter sensitivity of the steering stack.} Each block sweeps one axis with every other setting frozen at the final configuration, and best and second best are marked within each block. Macro averages on NegGrasp.}
\label{tab:hparam_sweep}
\end{table}

\paragraph{Number of particles $N$.}
Because $N$ is the population size of the particle filter rather than a selection budget, an $N$-particle run cannot be approximated by subsampling a larger one, so every row is generated from scratch. Constraint compliance improves monotonically with $N$ (violation 32.8\% to 17.2\%, contamination 15.5\% to 6.6\% from $N{=}2$ to $16$), CSR saturates between $N{=}8$ and $16$, and increasing to $N{=}32$ doubles inference cost without improving any metric, placing the default at the point where compliance has converged and further compute stops paying.

\paragraph{Weight sharpness $\lambda$.}
Raising $\lambda$ sharpens resampling toward the potential of Equation~\ref{eq:potential}, which rewards object proximity alongside avoidance, so violation drifts mildly upward as $\lambda$ grows (15.8\% at $\lambda{=}5$ to 19.5\% at $\lambda{=}50$), the same mechanism by which deterministic SMC reintroduces violations in Section~\ref{sec:ablation}. The drift stays small, with CSR confined to a 2.6-point band across the sweep, and the default is the only setting near the top of both the compliance and the physical columns.

\paragraph{Churn strength $\eta$.}
This is the axis the method is most sensitive to, and it quantifies the collapse mechanism of Section~4.4. At $\eta{=}0$ the sampler reduces exactly to the deterministic ODE and violation rises to 28.3\%, while every nonzero $\eta$ restores compliance to the 17-19\% range, confirming that the fix requires stochasticity but not a particular noise level. Because the churned update preserves the per-step marginals only up to discretization error, we verify preservation directly on a pilot protocol with guidance and resampling disabled, comparing each $\eta$ against its deterministic counterpart under pre-registered equivalence margins on compliance, success, and the particle dispersion of the grasp parameters. The check passes for $\eta \in \{0.05, 0.1\}$ but fails from $\eta{=}0.3$ upward, where the violation reduction exceeds the margin and is thus a distribution-distortion artifact rather than genuine avoidance. Within the admissible interval, $\eta{=}0.05$ is slightly stronger on CSR (65.9\%) while the default attains the lower violation rate (17.2\% against 19.3\%) at a nearly identical TSR, and beyond it the physical axis degrades steadily (TSR 38.7\%, 29.2\%, 27.0\% at $\eta{=}0.1$, $0.2$, $0.3$) with no compliance gain, as the injected noise pushes trajectories further from the learned flow.

\paragraph{Contact gate and guidance window.}
The two remaining CFG design choices are insensitive axes. Across a fivefold change of the gate distance (20 to 100\,mm), violation and CSR move by at most 2.5 and 1.3 points, consistent with the gate being self-limiting, since the guidance term vanishes for candidates already clear of the forbidden region regardless of the threshold. Restricting guidance to only the early or only the late half of the trajectory likewise changes compliance and CSR by about one point. Neither axis offers a setting that improves on the default, and neither degrades the method substantially, evidence that the contact-gated CFG design is robust rather than delicately tuned.

\paragraph{Potential saturation distances $\tau$ and $\delta$.}
The look-ahead potential of Equation~\ref{eq:potential} has two free parameters, the saturation distances of its avoidance and object-proximity terms, and both are low-sensitivity axes. Across a tenfold change of $\tau$ (8 to 80\,mm) and a fourfold change of $\delta$ (15 to 60\,mm), no setting improves on the default in either compliance or success, and CSR stays within 2.2 points of the default throughout, so the potential, like the contact gate and the guidance window, requires no per-task tuning.

\section{Compute Cost and Reproducibility}
\label{app:compute}

This appendix reports the compute cost of the steering stack and validates the stability and statistical significance of the reported numbers.

\paragraph{Inference cost.}
All experiments run on 4$\times$NVIDIA TITAN RTX (24\,GB) with the DiT forward pass in bfloat16 autocast. Generating, scoring, and simulating all 2{,}843 test instances takes about 1.6 hours per configuration end-to-end on this hardware, and the cost scales linearly with the particle count $N$.

\paragraph{Reproducibility.}
All experiments use a single fixed random seed (0), and every table is reproducible from the released per-case predictions and scoring code without re-running generation, as the scoring pipeline is deterministic. Generation itself is not bit-reproducible across processes (bfloat16 autocast and non-deterministic CUDA reductions), so we quantify stability rather than assume it: three independent runs of the final configuration give micro-averaged CSR of 70.2\%, 70.1\%, and 70.1\% ($\sigma{=}0.1$\,pp), while macro-averaged violation varies with $\sigma{=}2.3$\,pp, the same magnitude as re-running with an identical seed. We therefore treat this as run-to-run rather than seed variance and refrain from interpreting differences below these floors.

\paragraph{Statistical validation.}
To verify that the headline comparisons of Table~1 are not artifacts of the finite test set, we ran a stratified paired bootstrap ($B{=}10{,}000$, resampling cases within each of the 27 category-part cells and applying identical resample indices to all methods). Against the strongest baseline, DextER with best-of-$N$ selection, the 95\% intervals of the paired macro differences exclude zero on violation ($-12.6$ $[-17.8, -7.4]$), contamination ($-7.8$ $[-11.1, -4.3]$), and CSR ($+10.0$ $[+4.3, +15.7]$). The ablation claim of Section~4.4 is likewise confirmed, as churn reduces violation against deterministic SMC by $-11.1$ $[-16.1, -6.0]$.

%% file: Sections/appendix_model.tex
\section{Model}
\label{app:model}

This appendix gives implementation-level detail on the flow-matching DiT of Section~\ref{sec:method-arch}, which is trained once on DexGYSNet's positive instructions and left untouched by the inference-time steering of Sections~\ref{sec:method-cfg}--\ref{sec:method-smc}.

\textbf{Point cloud encoder.} A single frozen encoder, PartField~\cite{liu2025partfield}, supplies the geometric condition. PartField is a triplane-based encoder: a PVCNN backbone maps the point cloud into triplane features, and a triplane transformer refines these into $3{\times}32{\times}32{=}3{,}072$ patch tokens (three planes, each a $32 \times 32$ grid), which we average-pool $2{\times}2$ per plane down to $3{\times}16{\times}16{=}768$ patch tokens of width 1024. Input point clouds are sampled to 10{,}000 points (XYZ only; PartField's other three input channels are filled with a constant placeholder, matching its own pretraining pipeline). We load the released checkpoint (Hugging Face Hub, \texttt{mikaelaangel/partfield-ckpt}) and never fine-tune it, since fine-tuning measurably degrades its part-aware clustering (Figure~\ref{fig:partfield-degrade}).

\textbf{Text encoder.} A single frozen encoder, Qwen2.5-0.5B-Instruct~\cite{qwen2024report}, supplies the language condition. The positive instruction is wrapped in a fixed template, \texttt{"Grasp instruction: \{instruction\}"}, and encoded by the base transformer (not the language-modeling head); we keep every token of its last hidden state, $[1, T, 896]$, rather than pooling to a single vector, so the DiT's cross-attention can attend to individual words.

\textbf{Conditioning.} The two streams are independently projected (LayerNorm $+$ Linear) to a shared width of 768 and tagged with one of two learnable modality embeddings, then concatenated into a single memory of $T{+}768$ tokens that every DiT block cross-attends to; the two modalities are never summed or interleaved. Classifier-free guidance is realized architecturally: a single learnable \texttt{null\_text} token can replace the entire text stream (the geometry stream is never dropped, since the object being grasped should never be ambiguous). At training time this replacement is stochastic (each example's text condition is dropped with probability 0.15); at inference, Sections~\ref{sec:method-cfg}--\ref{sec:method-smc} instead replace it with the fixed second prompt $B$ rather than the null token, so the contrast used for steering is between two real instructions rather than between an instruction and nothing.

\textbf{State representation.} The denoised state $x = (x_g, x_c)$ (Eq.~\ref{eq:state}) is realized as 9 tokens: one grasp token, $x_g \in \mathbb{R}^{28}$ (3-D translation, 3-D axis-angle rotation, 22 ShadowHand joint angles) linearly projected to width 768 and given one learnable positional embedding, and 8 contact-slot tokens, $x_c \in \mathbb{R}^{8 \times 4}$ (each slot a 3-D coordinate plus a scalar occupancy flag) linearly projected and given 8 further learnable positional embeddings, one per slot.

\textbf{DiT architecture.} 12 blocks of width 768 with 12 attention heads and an MLP ratio of 4 ($3{,}072$-D hidden layer, GELU). Each block follows adaLN-zero: a sinusoidal timestep embedding (768-D, refined by a 2-layer SiLU MLP) is projected to 9 chunks of shift/scale/gate, one triple for each of the block's three sub-layers, self-attention, cross-attention, and MLP, and every gate is zero-initialized so each block starts as the identity. Self-attention is computed among the 9 state tokens alone; cross-attention queries those same 9 tokens against the concatenated text-and-geometry memory, masking only the padded positions of the (variable-length) text stream, since the 768 geometry tokens are never padded.

\textbf{Attention mechanism.} Every attention operation, both the self-attention among state tokens and the cross-attention into the conditioning memory, is fully bidirectional; no causal masking is used anywhere in the model. This is a consequence of the state being a small, fixed-size set of continuous tokens refined jointly by flow matching rather than a sequence emitted left to right, so there is no notion of "future" tokens to hide.

\textbf{Output heads and objective.} Three linear heads read out of the final block: a grasp velocity $\hat v_g \in \mathbb{R}^{28}$ from the grasp token, a contact velocity $\hat v_c \in \mathbb{R}^{8 \times 4}$ from the 8 slot tokens, and 18-way link-classification logits (17 ShadowHand link names $+$ one "none" class) from those same slot tokens. The two regression heads (grasp and contact velocity) are zero-initialized, so the model's continuous outputs start at exactly zero; the link-classification head keeps its default initialization, since a discrete classifier has no equivalent "start as identity" convention. Training regresses $\hat v_g$ against the rectified-flow target with an unweighted MSE, and regresses $\hat v_c$'s coordinate and occupancy channels against their own targets with two separate MSE terms (the coordinate term masked to slots the ground truth marks occupied), alongside a cross-entropy loss on the link logits; the four terms are combined with weights $1 : 1 : 1 : 0.5$ (grasp : coordinate : occupancy : link) in our released configuration.

%% file: Sections/appendix_qulitative.tex
\section{Qualitative Results}
\label{sec:qualitative}

\begin{figure*}[t]
\includegraphics[height=0.9\textheight,keepaspectratio]{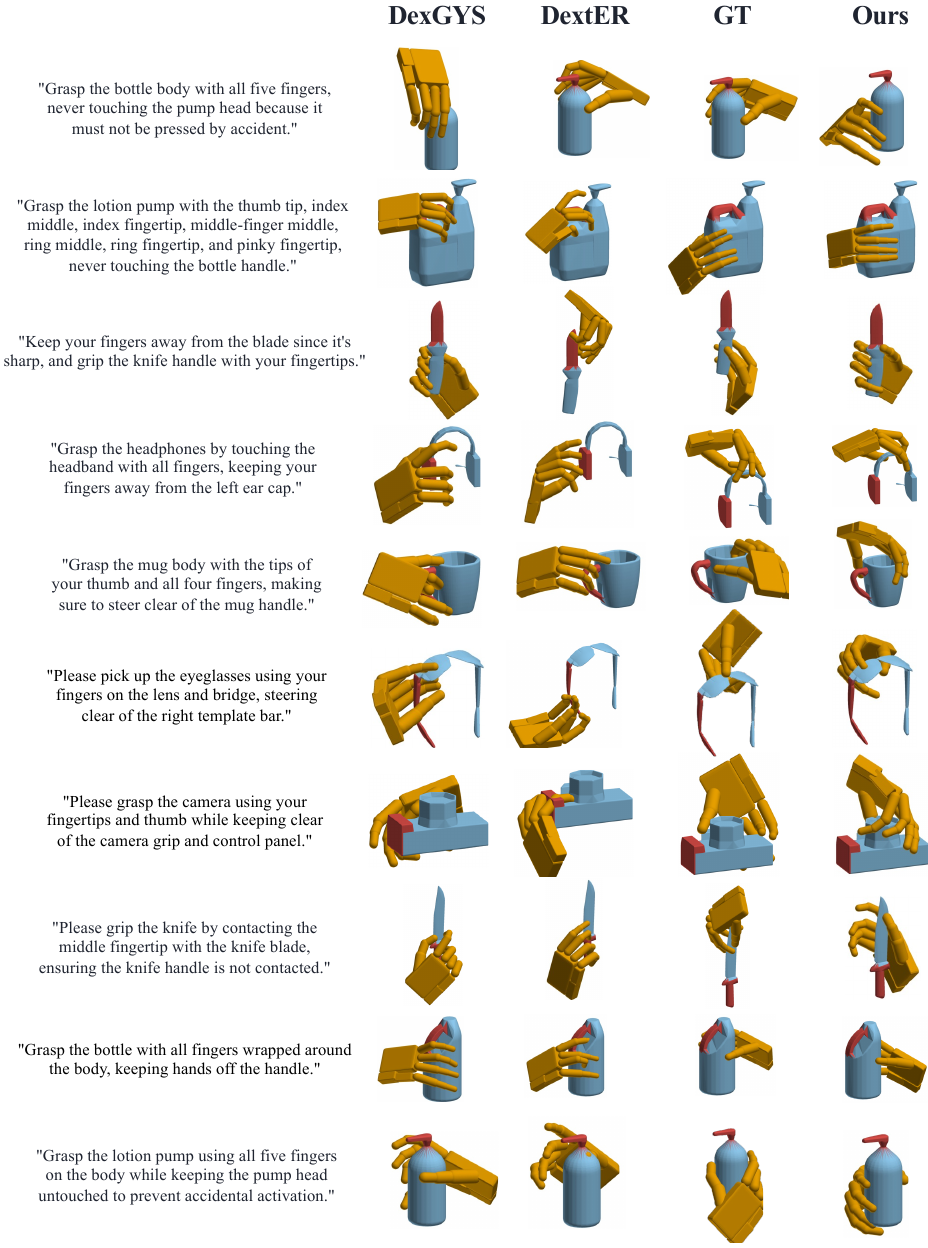}
\caption{\textbf{Qualitative comparison on NegGrasp.} The forbidden part is rendered in red. Baselines repeatedly contact or seize it, whereas our steering avoids it while reaching the instructed part.}
\label{fig:qualitative}
\end{figure*}

Figure~\ref{fig:qualitative} compares grasps on ten NegGrasp test cases spanning seven object categories, with the forbidden part rendered in red. The baselines frequently contact the forbidden part, and in several cases they seize it as if it were the instructed target, consistent with the shortcut hypothesis of Section~1 that a part mentioned in the instruction is read as a place to grasp. Our method stays clear of the forbidden part in every case shown while still contacting the instructed part, and its grasps often differ from the reference compliant grasp (GT), indicating that the steered sampler finds its own compliant mode rather than memorizing the reference.

\section{Limitation and Future Direction}
Our framework inherits the quality of its frozen grounding stack. When the parser or the 3D part-grounding model fails to localize the forbidden part, which occurs in 50 of the 2{,}843 generated test cases, steering has no region to act on, and since the parser keeps only the first flagged part, an instruction that forbids several regions at once is only partially enforced. Finally, all evaluation is conducted in simulation, on single-object scenes with a ShadowHand, and physical success is judged by the Isaac Gym protocol rather than on real hardware. Each of these limitations points to a natural extension, supporting multiple simultaneous constraints, improving open-vocabulary part grounding, and validating the steering stack on a physical robot in cluttered scenes, and because our method operates purely at inference time, every such extension leaves the trained generator untouched.